# CONTENT-BASED SPAM FILTERING ON VIDEO SHARING SOCIAL NETWORKS

*Antonio da Luz[1,2], Eduardo Valle[3], Arnaldo Araujo[1]*

[1] *NPDI Lab — DCC / UFMG, Belo Horizonte, MG, Brazil*
[2] *Federal Institute of Technology of Tocantins – IFTO, Paraíso, TO, Brazil*
[3] *RECOD Lab — IC / UNICAMP, Campinas, SP, Brazil*
*E-mail:* daluz@dcc.ufmg.br, mail@eduardovalle.com, arnaldo@dcc.ufmg.br

## ABSTRACT

In this work we are concerned with the detection of spam in video sharing social networks. Specifically, we investigate how much visual content-based analysis can aid in detecting spam in videos. This is a very challenging task, because of the high-level semantic concepts involved; of the assorted nature of social networks, preventing the use of constrained a priori information; and, what is paramount, of the context-dependent nature of spam. Content filtering for social networks is an increasingly demanded task: due to their popularity, the number of abuses also tends to increase, annoying the user base and disrupting their services. We systematically evaluate several approaches for processing the visual information: using static and dynamic (motion-aware) features, with and without considering the context, and with or without latent semantic analysis (LSA). Our experiments show that LSA is helpful, but taking the context into consideration is paramount. The whole scheme shows good results, showing the feasibility of the concept.

## 1. INTRODUCTION

In this paper, we are concerned with the detection of spam on video sharing social networks — online communities built upon the production, sharing and watching of short video clips, which have been nourished by the popularization of broadband web access and the availability of cheap video acquisition devices. The crowds of users who employ the services of websites like *Dailymotion*, *MetaCafe* and *YouTube*, not only post and watch videos, but also share ratings, comments, "favorite lists" and other personal appreciation data.

The emergence of those networks has created a demand for specialized tools, including mechanisms to control abuses and terms-of-use violations. Indeed, the success of social networks has been inevitably accompanied by the emergence of users with non-collaborative behavior, which prevents them from operating evenly. Those behaviors include instigating the anger of other users (*trolling*, in the web jargon), diffusing materials of genre inappropriate for the target community (e.g., diffusing advertisement or pornography in inadequate channels), or manipulating illegitimately popularity ratings.

Non-collaborative behavior pollutes the communication channels with unrelated information, and prevents the virtual communities from reaching their original goals of discussion, learning and entertainment. It alienates legitimate users and depreciates the social network value as a whole [1].

Specifically, we are interested on detecting spam on threads of video answers, a popular feature of some social networks, like *YouTube*, where the user can post a video in response to another. Here, we consider "spam" a video answer whose subject is unrelated with the original video (sometimes advertisement, commercial or not; sometimes videos posted in the hope to attract attention; sometimes videos posted intentionally to anger the other users). Figure 1 illustrates the diversity of the spam phenomenon.

## 2. RELATED WORK

The identification of pollution in online video social networks is a new topic with very few published works. After extensive research, we could only find [2] and [3]. The first tries to identify non-cooperative users and addresses both spamming and ballot stuffing (which they call "promoting"), by analyzing parameters like tags, user profile, the user posting behavior and the user social relations. The second classifies videos, accordingly to their pattern of access by users, into three categories: quality (the normal ones), viral (videos which experience a sudden surge in popularity) and junk (spurious videos, like spam).

Neither work use the video content itself for classification, instead, they rely on metadata and access logs. By contrast, our scheme mainly relies on the visual content.

Few works have been proposed to detect objectionable content in visual documents (images and videos). From those, the vast majority is concerned with nudity, pornography or graphical violence.

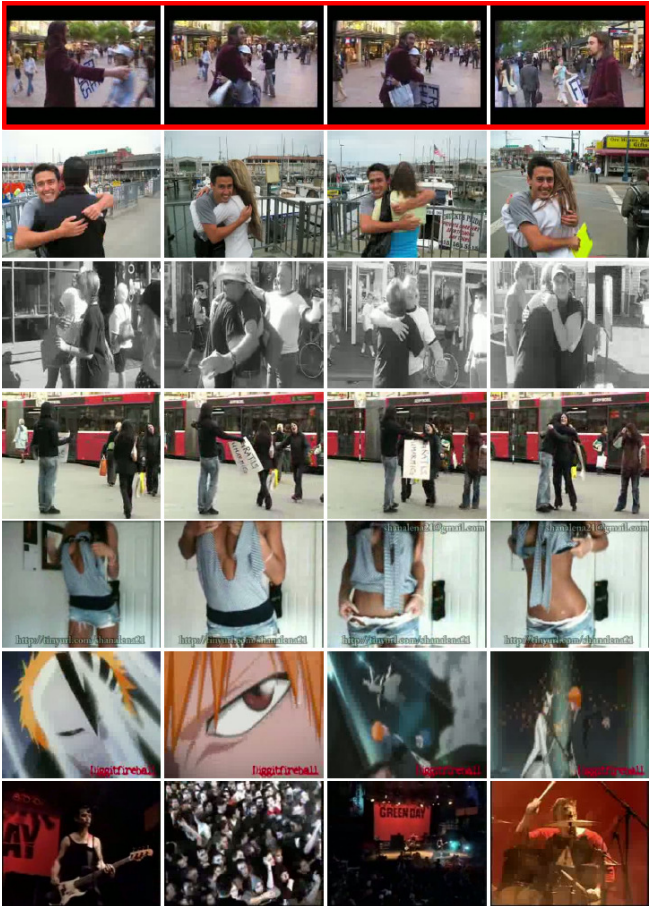

**Figure 1:** The complexity of *spam*. The first frames (topmost frames, red outline) are from the original video, which is related to the Free Hugs Campaign. The next three videos (one video per line) are *legitimate* ones. And, the last three are from *spam* videos.

The vast majority of pornographic detectors in images or videos is based on the detection of exposed skin [4] and seriously suffers from false positives of face close-ups, sport scenes or other innocent skin exposures. The available literature on violence detectors for content-filtering tends to concentrate on feature films and to give the soundtrack a very special attention (e.g. [5]). That kind of specialization warrants good performances, but makes the adaptation of those techniques difficult to the chaotic nature of social networks.

Recently, however, much attention has been devoted to less constrained approaches, using general-purpose features and classifiers. From those general approaches, one of most successful is the visual dictionary of local features.

The acceptance of local features as a broad technique of image description was an important watershed in the history of image understanding. Local features, like the popular SIFT descriptors [6], allow excellent discriminating power and great robustness to geometric and photometric transformations. If they were initially available only for static images, nowadays there exist local features that take into account the spatio-temporal nature of video, one of the most popular being STIP [7].

The discriminating power of local descriptors is extremely advantageous when matching objects in scenes, or retrieving specific target documents. However, when considering high-level semantic categories, it quickly becomes an obstacle, since the ability to generalize becomes then essential. A solution to this problem is to quantize the description spaces by using codebooks of local descriptors, in a technique sometimes named "visual dictionary". The visual dictionary is nothing more than a representation which splits the descriptor space into multiple regions, usually by employing non-supervised learning techniques, like clustering. Each region becomes then a "visual word" and is included in a "dictionary of visual words". The idea is that different regions of the description space will become associated to different semantic concepts, for example, parts of the human body, corners of furniture, vegetation, clear sky, clouds, features of buildings, etc. The technique has been employed successfully on several works for retrieval and classification of visual documents [8].

In addition to moderating the discriminating power of descriptors, the dictionaries allow adapting to visual documents techniques formerly available only to textual data. Among those borrowings, one of the most successful has been the technique of bags of words (which considers textual documents simply as sets of words, ignoring any inherent structure). The equivalent in the CBIR universe has been called bags of visual words, bags of features or bags of visual features, sometimes abbreviated as BoVF. It greatly simplifies document description, which becomes a histogram of the visual words it contains (Figure 2). The introduction of this technique had a huge impact on content-based retrieval and classification of visual documents [9].

The BoVF model also opens the opportunity to employ the Latent Semantic Analysis (LSA) [10].

The LSA is nothing more than an operation of change of basis in the document description in order to make more explicit some latent associations between them. Using the information provided by the bags of words, we create an occurrence matrix (telling which word occurs in which document, and usually applying some frequency normalization). LSA will then apply Singular Value Decomposition (SVD) to project this data in a new space of "topics". Usually the dimensionality of the topics space will be reduced, by discarding the component of low singular value. The documents can then be described by their histogram of topics instead of words.

LSA was initially intended for large corpora of text, but using the metaphor of the visual words has allowed employing it for visual documents. It has been applied on image tasks, achieving good results (e.g., [11] [12]).

## 3. DETECTING SPAM USING VISUAL CONTENT

Perhaps the most serious problem to detect spam in video threads is its relative, context-dependent, nature. Accepting the definition that a spam video is simply a video unrelated to the thread topic, the same video (e.g., a viral video with a celebrity breakdown) may be spam in a thread (e.g., a thread about how to cook asparagus correctly), but not in another (e.g., a thread about famous people behaving oddly). It is, of course, possible that some videos are intrinsically more probable to be used as spam (like viral videos) than others (like someone cooking asparagus), but this does not solve, by itself, the problem.

The other serious difficulty is the large variety of visual content that can be found in legitimate elements. Even considering a very restricted thread (e.g., "how to cook asparagus") and observing only the legitimate answers, the diversity of the videos is overwhelming. Even a human operator has sometimes difficulty in establishing the legitimacy × spam status of the videos by watching just the images.

Finally, we face other difficulties, dictated by the flexible nature of the social networks. The number of videos in the threads is not fixed, and is usually quite small. Examples of spam are uncommon and tend to appear in clusters. This is fortunate, in one way, but it makes training the classifier quite difficult.

To test those hypotheses, we have built a video classification system, based on the following scheme:

1. The classes considered were spam (positive) and legitimate (negative);
2. We start by extracting the low level features. Those were either STIP [7] features extracted from the videos, either SIFT [6] features extracted from keyframes of the video (the keyframes were extracted using a state-of-the-art static video summarization method [13]).
3. The codebook (visual dictionary) was constructed by choosing at random 5000 low level features.
4. Using the codebook, each video was given a description in terms of a bag-of-visual-features (BoVF), which is simply a histogram of the low-level features, quantized by their proximity to the codebook;
5. As an optional step, the BoVF is projected in the latent semantic space was LSA (preserving all topics with non-zero singular values);
6. Also as an optional step, context information can be incorporated by making the description relative to the original videos. This is made simply by taking the video feature vector (either in the BoVF or topic space) and subtracting the feature vector of its corresponding original video;
7. To classify the videos, we have used the well-known SVM with linear kernel.

The idea of step 5 is to detect cross-video appearance correlations which might be revealing of semantic patterns.

In a nutshell, while the BoVF-model takes the appearances at "face value", the Visual Topics-model of LSA looks for hidden patterns that may indicate that different visual words may be related or that certain combinations of visual words may make sense together. That might alleviate the problem of extreme visual diversity in legitimate and spam patterns.

Step 6 tries to solve the problem of context, creating bags of "differences" between the answer video and the original video (those differences can be between words or topics). Abstractly, the context-free video description was a vector in the BoVF space or the topic space. The context-aware description is now the vector difference between the video and its original one. This allows us to take the different contexts (the different threads) while keeping the classification model extremely simple (only two classes, spam and legitimate, for the entire dataset). As far as we know, this use of "bags of differences" as a context-aware description model is original.

## 4. EXPERIMENTS AND RESULTS

Given the novelty of this application, it is unsurprising that no standard database is available for evaluation purposes. Therefore, we have collected ourselves 8182 videos from 84 threads, chosen at random from the "Most Responded Videos" list generated by YouTube. The collection took place in mid-2008. We have selected the videos in the "most responded" list, because they form long threads, often with a lot of spam.

From the 8182 videos, 84 were the original "heads of the thread". From the remainder answers, manual inspection determined 3420 to be legitimate and 4678 to be spam. Deciding which videos are spam is sometimes hard, even for humans: we have considered them as spam when their visual contents did not match the subject of the thread. In case of doubt, we adopted the policy of [3] and marked the video as legitimate.

For the experiment reported here, we have subsampled the original dataset, randomly selecting (besides the 84 originals) 1000 legitimate and 1000 spam.

The experimental design was a classical 5-fold cross-validation, generating approximately 1600 videos for training and 400 for testing on each fold. The numbers reported are the average of the folds.

Figure 4 is presented the results with the evaluation of the two purposed approaches, with both visual features, and a baseline result in an experiment without considers the context location of the videos (traditional BoVF).

The results indicate that the context information is critical to identifying spam using a two-class classification model. The visual characteristics of the video allows filtering some of the spam (blue circles in Figure 4): this is interesting and show that, at least in our sample, videos used as spam tended to share some visual characteristics.

However, even the worst context-aware experiment (red data points) was best than the best context-blind experiment.

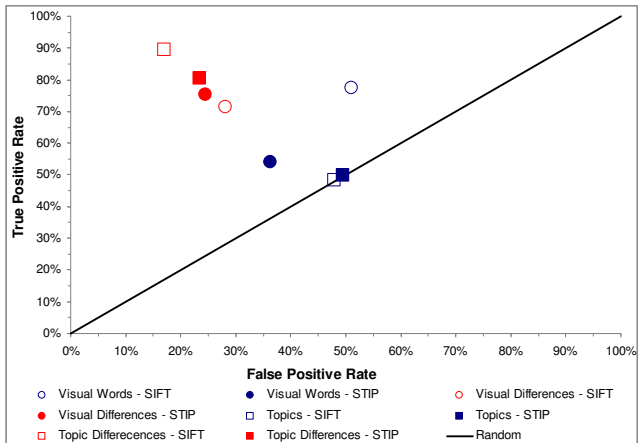

**Figure 4**: Experimental results. The sweet spot is the upper left corner. The data points represent different choices: empty × filled symbols are SIFT × STIP; circles × squares are BoVF × topics; blue × red are without and with context. The use of context is critical (the worst experiment with context is still better than the best without context). In this case, the use of LSA (to project BoVF on topic space) is best (red squares).

The use of LSA to pass from the BoVF space to the topic space proved advantageous, especially when using SIFT descriptors, which, in association with context information have, in the topic space, the best performance overall.

The choice of low-level feature is somewhat inconclusive. For the context-blind experiments, they present mainly a different in bias (with SIFT tending to have more false positives, and STIP more false negatives). For the context-aware experiments, STIP was better in the BoVF space, and SIFT was better in the topic space. SIFT tends to generate more local features than STIP: it is thus possible that the difference observed is a consequence of SIFT being denser.

## 5. DISCUSSION

Removing content automatically is only possible when false positive rates are very low, because removing a legitimate answer is much more problematic than accepting a spurious one. At this stage, our technique, used in isolation, does not allow such low rates and thus cannot be used to forceful removing content from the social network. That does not mean, however, that the technique is of no practical value. An interesting strategy may be employed to make it feasible: combining it with manual inspection of the "suspect" videos, in order to only remove those which are indeed deemed as illegitimate.

Of course, our current approach explores only the *visual* information contained in the video, and thus is only the lower bound on what could be obtained adding other evidences, such as those provided by the soundtrack, metadata, social network statistics, etc. Currently we are investigation on how to best incorporate our visual classifier in a system taking into account all those evidences.

## 6. ACKNOWLEDGMENTS

The authors are thankful to the Brazilian agencies CNPq, CAPES, FAPEMIG and FAPESP, for the financial support.

## 7. REFERENCES

[1] T. Deselaers, L. Pimenidis and H. Ney. "Bag-of-Visual-Words Models for Adult Image Classification and Filtering", In: *International Conference on Pattern Recognition*, pp. 1-4, 2008.

[2] F. Benevenuto, T. Rodrigues, V. Almeida, J. Almeida and M. Gonçalves, "Detecting Spammers and Content Promoters in Online Video Social Networks", In: *International ACM SIGIR Conference on Research and Development in Information Retrieval*, pp. 620-627, 2009.

[3] R. Crane and D. Sornette, "Robust Dynamic Classes Revealed by Measuring the Response Function of a Social System", In: *National Academy of Sciences*, 105(41):15649-15653, 2008.

[4] W. Kelly, A. Donnellan, D. Molloy. "Screening for Objectionable Images: A Review of Skin Detection Techniques", In: *International Machine Vision and Image Processing Conference*, pp. 151-158, 2008.

[5] T. Giannakopoulos, D. I. Kosmopoulos, A. Aristidou, and S. Theodoridis, "Violence Content Classification Using Audio features", In: Hellenic Artificial Intelligence Conference SETN-06, LNAI 3955, pp. 502-507, 2006.

[6] D. G. Lowe. "Distinctive Image Features from Scale-Invariant Keypoints", In: *International Journal of Computer Vision*, vol. 60, no. 2, pp. 91-110, 2004.

[7] I. Laptev. "On Space-Time Interest Points", In: *International Journal of Computer Vision*, vol 64, number 2/3, p.107-123, 2005.

[8] J. Sivic and A. Zisserman. "Video Google: A Text Retrieval Approach to Object Matching in Videos", In: *IEEE International Conference on Computer Vision*, pp. 1470-1477, 2003.

[9] J. Yang, Y.-G. Jiang, A. G. Hauptmann and C.-W. Ngo. "Evaluating Bag-of-Visual-Words Representations in Scene Classification", In: *ACM MIR'07*, pp. 197-206, 2007.

[10] T. Landauer, P. Foltz and D. Laham. "Introduction to Latent Semantic Analysis". *Discourse Processes*, 25, 259-284, 1998.

[11] J. C. Caicedo, J. G. Moreno, E. A. Niño, and F. A. Gonzalez. "Combining visual features and text data for medical image retrieval using latent semantic kernels". In *ACM MIR'10*. ACM, New York, NY, USA, 359-366, 2010.

[12] K. Yanai and K. Barnard. "Region-based automatic web image selection". In *ACM MIR'2010*. ACM, New York, NY, USA, 305-312, 2010.

[13] S. Avila, A. Lopes, A. da Luz Jr., and A. de A. Araujo. "VSUMM: A mechanism designed to produce static video summaries and a novel evaluation method". *Pattern Recogn. Lett*. 32, 1, 56-68, 2011.